# AI Magnetic Levitation (Maglev) Conveyor for Automated Assembly Production

Ray Wai Man Kong[1]

[1] Adjunct Professor, City University of Hong Kong, Hong Kong

[1] Modernization Director, Eagle Nice (International) Holding Ltd, Hong Kong

***Abstract:*** **Efficiency, speed, and precision are essential in modern manufacturing. AI Maglev Conveyor system, combining magnetic levitation (maglev) technology with artificial intelligence (AI), revolutionizes automated production processes. This system reduces maintenance costs and downtime by eliminating friction, enhancing operational efficiency. It transports goods swiftly with minimal energy consumption, optimizing resource use and supporting sustainability. AI integration enables real-time monitoring and adaptive control, allowing businesses to respond to production demand fluctuations and streamline supply chain operations.**

**The AI Maglev Conveyor offers smooth, silent operation, accommodating diverse product types and sizes for flexible manufacturing without extensive reconfiguration. AI algorithms optimize routing, reduce cycle times, and improve throughput, creating an agile production line adaptable to market changes.**

**This applied research paper introduces the Maglev Conveyor system, featuring an electromagnetic controller and multiple movers to enhance automation. It offers cost savings as an alternative to setups using six-axis robots or linear motors, with precise adjustments for robotic arm loading. Operating at high speeds minimizes treatment time for delicate components while maintaining precision. Its adaptable design accommodates various materials, facilitating integration of processing stations alongside electronic product assembly. Positioned between linear-axis and robotic systems in cost, the Maglev Conveyor is ideal for flat parts requiring minimal travel, transforming production efficiency across industries. It explores its technical advantages, flexibility, cost reductions, and overall benefits.**

## I. INTRODUCTION

In the rapidly evolving landscape of modern manufacturing, the pursuit of efficiency, speed, and precision has become paramount. As businesses strive to enhance their production capabilities, innovative technologies are emerging to meet these demands. One such groundbreaking advancement is the AI Maglev Conveyor system, which leverages magnetic levitation (maglev) technology combined with artificial intelligence (AI) to revolutionize automated production processes.

From a business perspective, the AI Maglev Conveyor and Maglev Conveyor offer a transformative solution that addresses the challenges of traditional conveyor systems. This technology significantly reduces maintenance costs and downtime by eliminating friction and wear associated with conventional conveyors, leading to enhanced operational efficiency. The ability to transport goods at high speeds with minimal energy consumption not only optimizes resource utilization but also contributes to a more sustainable manufacturing environment. Furthermore, the integration of AI allows for real-time monitoring and adaptive control, enabling businesses to respond swiftly to fluctuations in production demands and streamline their supply chain operations.

Efficiency is at the core of the AI Maglev Conveyor's design. The system's magnetic levitation mechanism allows for smooth and silent operation, minimizing disruptions in the production environment. With the capability to handle diverse product types and sizes, the AI Maglev Conveyor enhances flexibility in manufacturing processes, accommodating varying production requirements without the need for extensive reconfiguration. Additionally, AI algorithms can analyze data from the conveyor system to optimize routing, reduce cycle times, and improve overall throughput. This results in a more agile production line that can quickly adapt to changing market conditions and customer demands.

### *A. Business Trend in the Marketing Research*

Referred to the DiMarket research report [1], the Planar Maglev Conveyor represents a significant leap forward in automated production technology. By combining the advantages of magnetic levitation with the intelligence of AI, this innovative system not only enhances business efficiency and reduces operational costs but also positions manufacturers to thrive in an increasingly competitive landscape. As industries continue to embrace automation, the AI Maglev Conveyor stands out as a key enabler of future-ready production environments.

The Magnetic Levitation (Maglev) Conveyor System market is experiencing significant growth, driven by increasing automation demands across various sectors. Key advantages of this technology include high speed, precision, reduced maintenance needs, and enhanced efficiency compared to traditional conveyor systems. Major applications are found in the food and beverage industry, electronics, medical sectors, and the automotive industry, all of which benefit from maglev's capabilities. Although the initial investment is higher than traditional systems, long-term savings from reduced downtime and increased throughput justify the cost.

The market is segmented by application (food and beverage, industrial, medical, electronics, automotive, and others) and type (single-track and double-track), with double-track systems gaining traction in high-volume applications. North America and Europe currently dominate the market due to early adoption and advanced automation infrastructure, while rapid industrialization in the Asia Pacific, particularly in China and India, is expected to drive future growth.

The market is projected to grow at a Compound Annual Growth Rate (CAGR) of 15% from 2025 to 2033, despite challenges such as high initial costs and the need for specialized expertise. The Magnetic Levitation (Maglev) Conveyor System market has shown the sales projection growth from 225% in the year 2025-2026 and ramp up to 600% in the year 2032- 2033 in Fig. 1.

However, ongoing technological advancements and rising labour costs are helping to mitigate these issues. Leading companies like Bosch Rexroth, Beckhoff, and LS Electric are driving innovation and expanding their presence through product development and strategic partnerships. Overall, the future of the Planar Maglev Conveyor System market looks promising, with continued growth and expanding applications across various industries globally.

The planar Maglev conveyor system market is experiencing robust growth, driven by increasing automation needs across various industries. From 2019 to 2024 (historical period), the market witnessed steady growth, and this trend is projected to accelerate during the forecast period (2025-2033). The estimated market value in 2025 is in the hundreds of millions of USD, and this figure is expected to reach several billion USD by 2033. This growth is fueled by several factors: the rising demand for high-speed, high-precision material handling in industries such as electronics and automotive; the increasing adoption of Industry 4.0 technologies and smart factories; and continuous advancements in Maglev technology, leading to improved efficiency, reliability, and affordability.

Furthermore, the growing need for hygienic and clean transportation solutions in industries like food and beverage is bolstering the market's expansion. The shift toward sustainable manufacturing practices is also contributing to the adoption of energy-efficient Maglev systems. Competition is expected to intensify, with companies focusing on product differentiation through advanced features and customized solutions. The development of new materials and technologies to enhance durability and energy efficiency will be key to future market growth. The growing adoption of automation in emerging economies is also a significant contributor to this expansion. Finally, governmental incentives and supportive policies aimed at boosting industrial automation are furthering market growth.

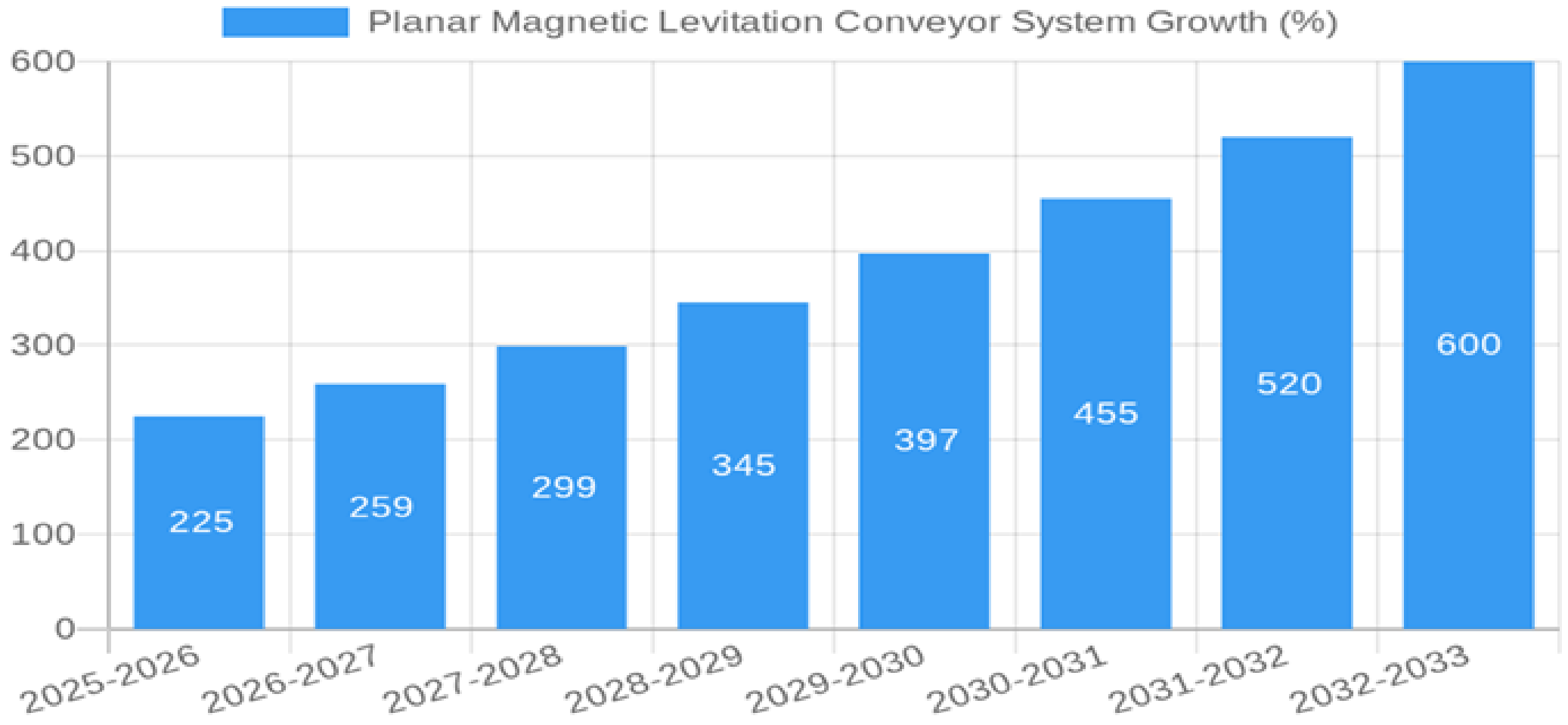

**Figure 1: Planar Magnetic Levitation Conveyor System Growth (%)**

### *B. Benefits of Magnetic Levitation Conveyor and its system*

Magnetic Levitation (Maglev) Conveyor systems offer several significant benefits that make them an attractive option for various industries. Here are some of the key advantages:

- High Speed and Efficiency

Rapid Transport: Maglev conveyors can achieve high speeds due to the lack of friction between the conveyor and the transported items. This allows for faster movement of goods, which is crucial in high-demand environments.

Increased Throughput: The ability to move items quickly and efficiently can lead to higher production rates and improved overall throughput in manufacturing processes.

Reduced Maintenance Needs

Minimal Wear and Tear: Since maglev systems operate without physical contact between moving parts, there is significantly less wear and tear compared to traditional conveyor systems. This results in lower maintenance costs and fewer frequent repairs.

Longer Lifespan: The reduced friction and wear contribute to a longer operational lifespan for the conveyor system.

- Precision Handling

Accurate Positioning: Maglev conveyors provide precise control over the movement of items, making them ideal for applications that require exact positioning, such as in the electronics and medical industries.

Gentle Handling: The smooth operation minimizes the risk of damage to delicate products, ensuring that items are transported safely.

- Energy Efficiency

Lower Energy Consumption: Maglev systems typically consume less energy than traditional conveyors due to their frictionless operation, leading to cost savings on energy bills and a reduced environmental footprint.

- Flexibility and Scalability

Modular Design: Maglev conveyor systems can be designed in modular configurations, allowing for easy expansion or reconfiguration as production needs change.

Adaptability: They can handle a variety of product sizes and shapes, making them versatile for different applications.

- Hygienic Operation

Clean Environment: The absence of physical contact reduces the accumulation of dust and debris, making maglev conveyors suitable for industries with strict hygiene requirements, such as food and beverage processing.

- Noise Reduction

Quiet Operation: Maglev systems operate quietly due to the lack of mechanical friction, contributing to a more pleasant working environment and reducing noise pollution in manufacturing facilities.

- Enhanced Safety

Reduced Risk of Accidents: The smooth and controlled movement of items minimizes the risk of accidents associated with traditional conveyor systems, improving workplace safety.

Overall, Magnetic Levitation (Maglev) Conveyor systems provide a range of benefits that enhance operational efficiency, reduce costs, and improve product handling across various industries. Their unique technology positions them as a forward-thinking solution for modern automated production environments.

## II. LITERATURE REVIEW FOR DESIGN OF MAGNETIC LEVITATION (MAGLEV) CONVEYOR

Referring to the Analysis of the Influence of Magnetic Field in the Propulsion Force of a Permanent Magnet-HTS Hybrid Magnetically Levitated Transport System from Takinami, A. H., Maeda, A., Abe, Y., & Ohashi, S. [2], the Fig. 2 and 3 show the experimental device in which all permanent magnets are made of neodymium magnets, including the magnet rails that are magnetized uniformly in the propulsion direction, and show the propulsion coils installed above the rail.

The transporter achieves levitation above the magnetic rail due to the high-temperature superconductor (HTS)'s pinning force, enabling unimpeded movement along the rail without the need for gap control. Since carrying weight requires a large levitation force, permanent magnets are installed under the load stage to take advantage of the repulsive force between them and the magnetic rail.

Moreover, the load stages are affixed to the transporter through a linear guideway, affording it independent vertical mobility and relying exclusively on repulsive permanent magnets for support. Fig. 3 shows the propulsion force measuring method using the Two-coil method, where the magnetic gradient is generated by demagnetizing the coils under the HTS front side and magnetizing the coil under the rear side of the HTS. Fig. 4 shows the propulsion method using the One-coil method, consisting of magnetizing the coil just under the HTS. In both methods, the HTS bulks tend to tilt to create forward momentum, and these are used for both levitation and propulsion. In addition, the Yoke iron in both Fig. 3 and Fig. 4 has been omitted for simplification.

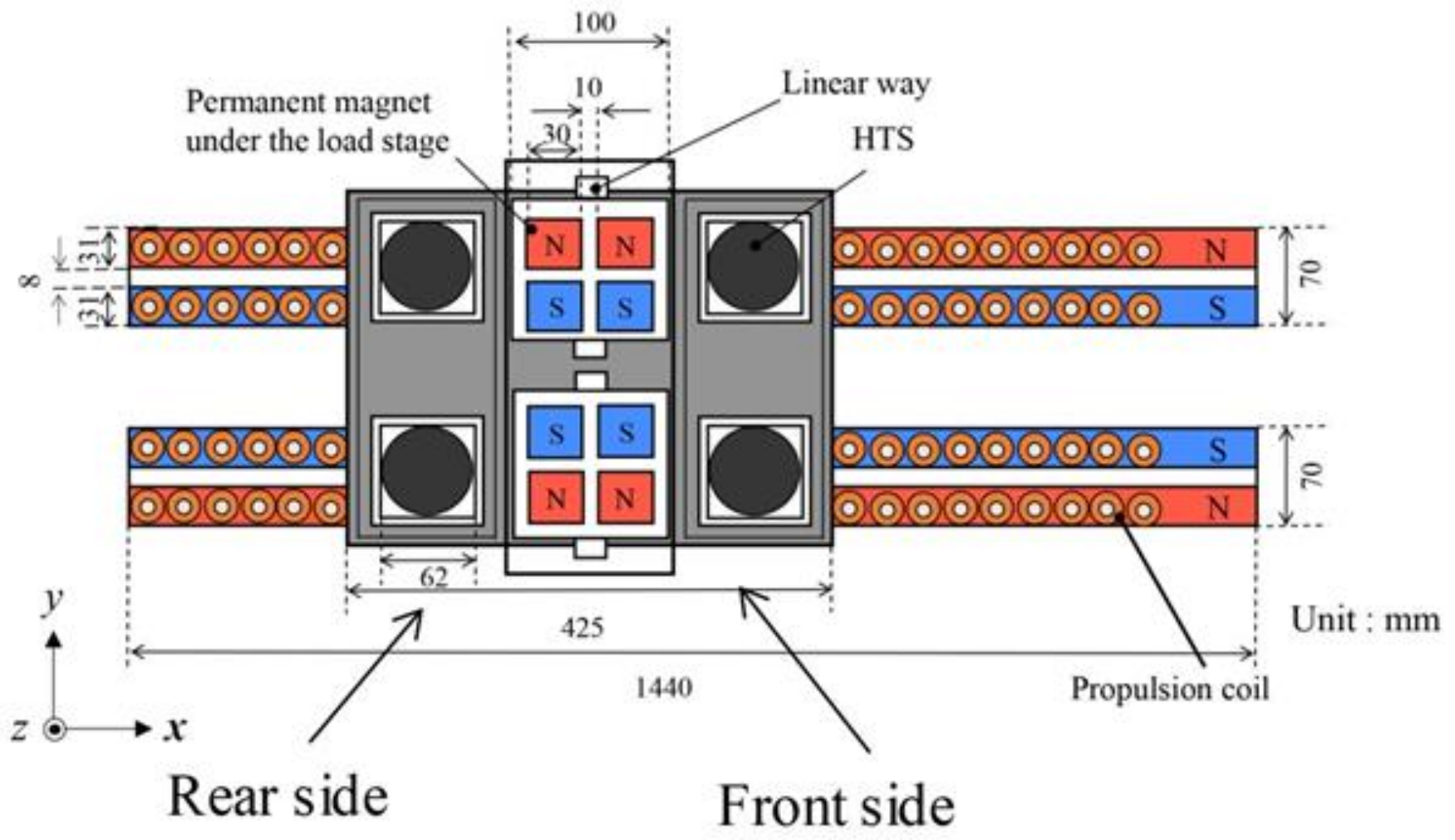

**Figure 2: Schematic Diagram of Maglev Conveyor Construction from Takinami et al.**

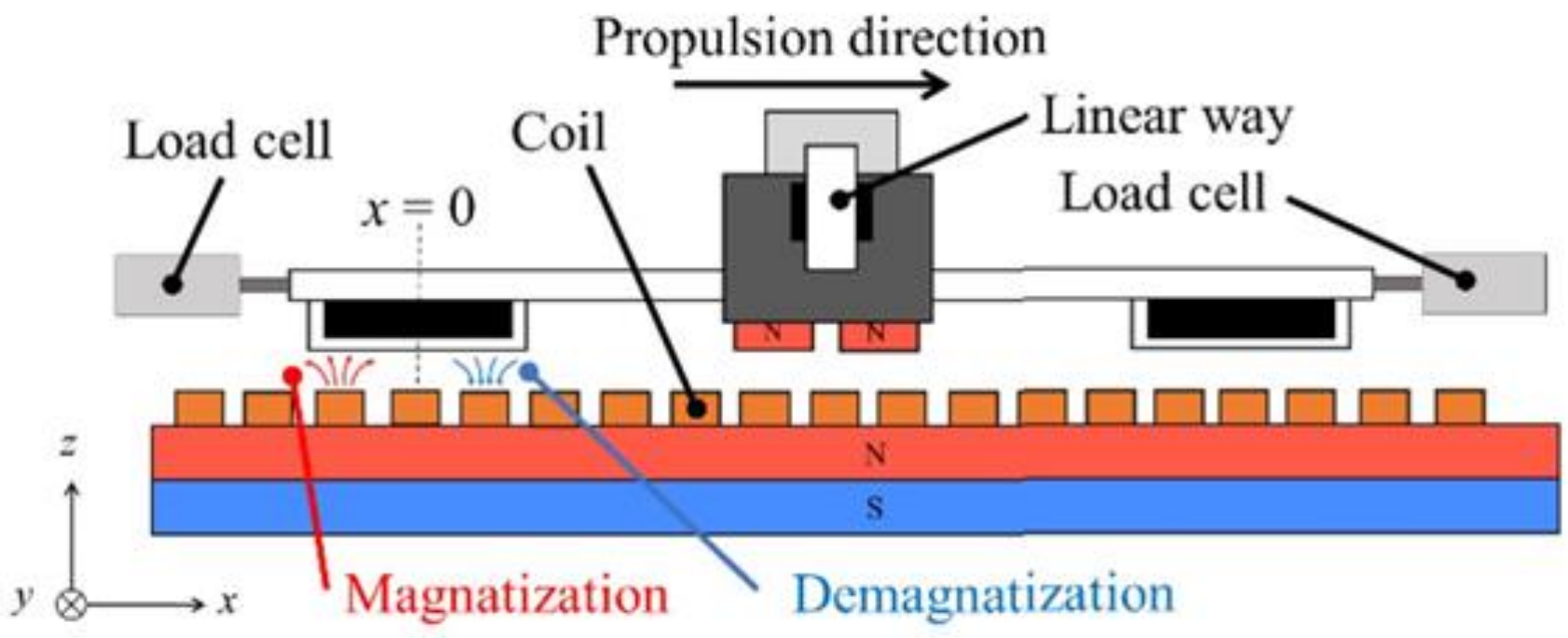

**Figure 3: Propulsion force measuring using the two-coil method from Takinami et al.**

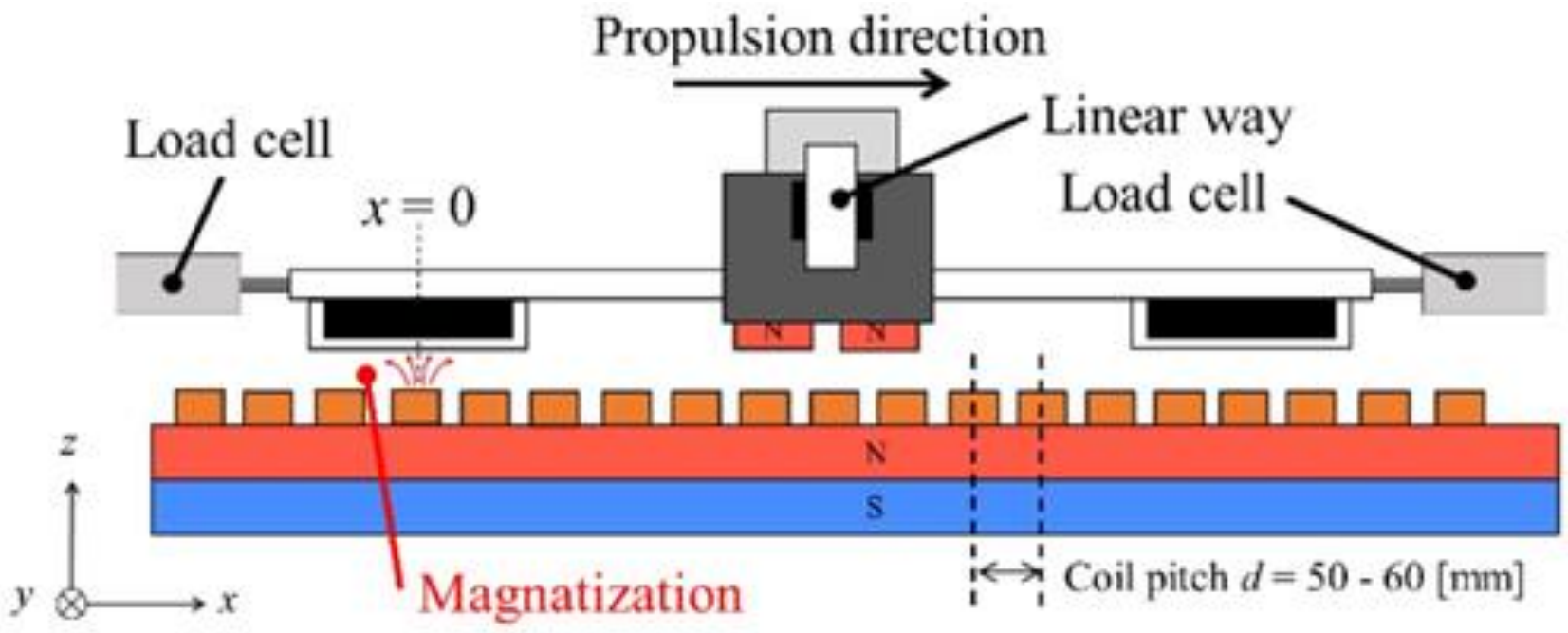

**Figure 4: Propulsion force measuring using the one-coil method from Takinami et al.**

Hence, the larger the inductance, the larger the flux density variation rate. Thus, the experimental device has been developed, and for the one-coil method experiment, an excitation current I = 1, 3 A was used, and for the two-coil method, I = 3 A was used, in which each coil is excited with 1.5 A. Four HTS bulks were installed on the transporter and cooled in containers filled with liquid nitrogen under the influence of the flux of the magnetic rail (field cooling) at the desired gap z = 23 mm, subsequently, levitating the transporter without control. The flux is measured by the hall sensor, and the propulsion force (static force) is measured by using load cells, two at the front and the other two at the rear side of the transporter.

The basic system for 1DOF movement consists of two serpentine traces, individually actuated from Jiri, Kuthan et al[3], Juřík, Martin et al. [4], as referring to the Magnetic levitation in Wikipedia [5]. Fig. 5 shows the schematic of the trace paths and a 3x3 magnet microrobot on top. On position number 1, the magnets are in their equilibrium position where the magnetic flux density is the highest, between two opposite currents from the same trace path.

On moving from 1 to 2, the first trace path is turned off while the second is turned on. This causes the magnets to move to their new equilibrium, toward the higher magnetic flux density.

Repeating this procedure with opposite currents on the same trace paths, a movement in the desired direction is produced by Pelrine, R et al [6].

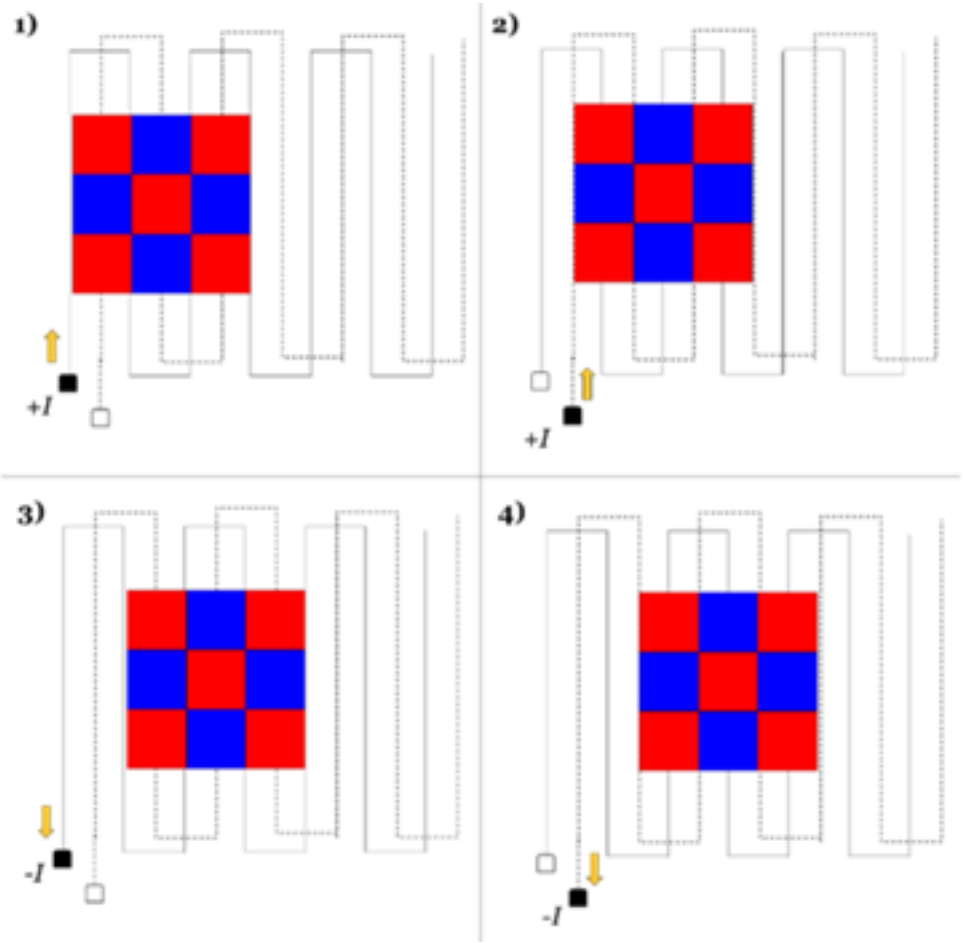

**Figure 5: Position transitions of a levitating microrobot using two pairs of serpentine traces**

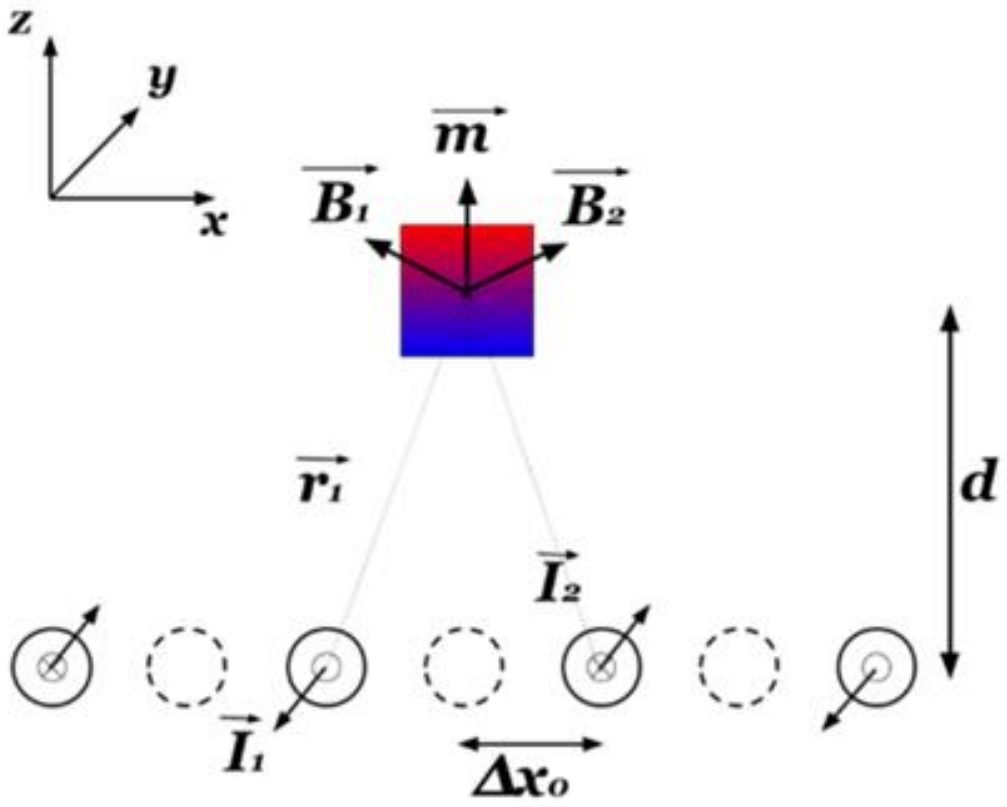

**Figure 6: Magnetic flux vectors induced on a magnet cube over a pair of serpentine traces**

To find the velocity, the forces on the microrobot must be analysed in Fig. 6. The microrobot is supposed to levitate, and so no friction force is produced, other than the air drag, which is also not considered.

The force produced by the interaction of the magnetic moments of the microrobot and the flux density of the serpentine traces is:

$$F = \nabla\, (\vec{m} \,.\, \vec{B}) \qquad (1)$$

Where F is a force, $\vec{m}$ is the magnetic moment that can be applied to a planar mover, $\vec{B}$ is the magnetic field.

The magnetic moment vector, given the orientation requirement for the diamagnetic levitation, is:

$$\vec{m} = (0 \;\; 0 \;\; \frac{B_r\, V_m}{\mu_0}) \qquad (2)$$

Meanwhile, the contribution to the $B$ field by the 2 closest traces in Fig. 6 is:

$$\vec{B} = \vec{B}_1 + \vec{B}_2 = \mu_0 \frac{\vec{I}_1 \vec{r}_1}{2\pi |\vec{r}_1|^2} + \mu_0 \frac{\vec{I}_2 \vec{r}_2}{2\pi |\vec{r}_2|^2} \quad (3)$$

Where B

Since for this approximation $\vec{m} . \vec{B}$ is not dependent on y or z, their derivatives are zero, and only force in the x direction is produced:

$$F_x = \frac{B_r V_m I}{2\pi} \frac{\delta}{\delta_x} \left[ \frac{x - \Delta x_0}{(x - \Delta x_0)^2 + d^2} + \frac{x + \Delta x_0}{(x + \Delta x_0)^2 + d^2} \right] \quad (4)$$

Where $F_x$ is a force in the x-axis, $B_r$ is a magnetic field related to $r$, $V_m$ *is* the velocity of a moving charge of magnetic, $I$ is the current, $\Delta x_0$ is the change of location in the x-axis, $d$ is the distance between the magnetic and the coil.

This is the only force applied to the magnet, and it can be equated to the robot's mass multiplied by its acceleration. This equation can be integrated to find the velocity of the microrobot:

$$F_x = mv \frac{d_v}{d_x} \quad (5)$$

$$F_x = \frac{B_r V_m I}{2\pi m} \left( \frac{2\Delta x_0}{(4\Delta x_0)^2 + d^2} \right) = \frac{1}{2} v_{max}^2 \quad (6)$$

Introducing the relation between magnet volume $V_m$, mass $V_m$, and density$\rho_m$, The related formula is $V_m = \frac{m}{\rho_m}$, in the previous equation the mass cancels out, which means that if more magnets are added (N number of magnets), the force will increase linearly.

It shows the force on the x-axis is moving the object in the x direction.

## III. DESIGN OF MAGNETIC LEVITATION CONVEYOR

***Case Study of Maglev Conveyor Design in Company Y***

The design of the Maglev Conveyor in Company Y has included research on the magnetic principle of the Maglev Conveyor. In the magnetic levitation system, electromagnetic force is the key to achieving suspension and drive. For electromagnetic levitation, according to Ampere's law and Biot-Savart law, the magnetic induction intensity B.

$B = \frac{\mu_0 IN}{2R}$ . The $B$ is generated by the magnetic field. It is generated by a single energized coil at a distance $R$, it can be expressed as:

$$B(z) = \frac{\mu_0 IN}{2R} \quad (7)$$

Where $\mu_0$ is the magnetic permeability of the vacuum ($\mu_0 = 4\pi * 10^{-7} H/m$). $I$ is the current in the coil, $N$ is the number of turns in the coil, and $R$ is the radius of the coil.

When a suspended object (with magnetic components) is in the magnetic field, the electromagnetic force $F$ received can be calculated by the Lorentz force formula. $F = qvBsin\theta$ (Macroscopic objects can be calculated using equivalent magnetic charge and other methods. In common cases of magnetic levitation, if the magnetic field is perpendicular to the magnetic direction of the object, etc., it is simplified. For the force on a magnetic body in a magnetic field, it can also be calculated by the energy method. For example, based on the magnetic energy density $\mu_m = \frac{B^2}{2\mu_0}$. The force is obtained by differentiating the magnetic energy concerning the displacement.

Taking a simple parallel plate magnet model as an example, the electromagnetic force $F$ is related to the magnetic flux $\emptyset$, the air gap length $g$, etc.:

$$F = -\frac{\partial W_m}{\partial g} \quad (8)$$

Where $W_m$ , is magnetic energy, $W_m = \frac{1}{2}LI^2$ (L is inductance). In this case, the inductance can be simplified by the magnetic circuit method to obtain the electromagnetic force.

**Kinematic equation**

The movement of the magnetic levitation conveyor follows Newton's laws of motion. Assume that the conveyor load mass is *m*, and the total external force (including electromagnetic force $F_{em}$, friction force $F_f$, etc., generally, magnetic levitation can greatly reduce friction, $F_B$ . There is a friction from the electromagnetic resistance, and it can be ignored in ideal conditions. It satisfies its acceleration *a* , as below:

$$F_{net} = F_{em} - F_f - F_B = ma \quad (9)$$

In the case of linear motion, if the initial velocity is $v_0$. The velocity *v* and displacement *x* after time *t* are shown below:

$$v = v_0 + at \quad (10)$$

$$x = v_0 + \frac{1}{2}at^2$$

From Zhang, Wenbai et. al.[7], the tractive force of the maglev train is provided by the long stator synchronous linear motor; *ψd, ψq, id, iq* are the magnetic linkage and current components of the stator winding on the *d-q* axis, respectively; and τ is the stator pole distance. According to the rotor magnetic field-oriented control strategy, *id* is usually controlled to 0, and the model is described as follows:

$$F = \frac{3\pi}{2\tau}(\psi_d\ i_q - \psi_q\ i_d) \quad (11)$$

when $i_d = 0$ . Then the tractive force is shown, $F = \frac{3\pi}{2\tau}(\psi_d\ i_q)$

Company Y has developed the magnetic levitation conveyor for automation. The automation is required to have a high precision of jig and fixture to locate the conveyor for the automatic pick and place process, checking process, and other processes. Prof Dr Ray Wai Man Kong [8] [9] [10] [11] [12] [13] [14] [15] [16]has used the lean methodology and conducted more applied research on an automatic gripper for the manufacturing process to be automated. Fig. 7 shows the schematic diagram of the design of the magnetic levitation conveyor for automation from Company Y.

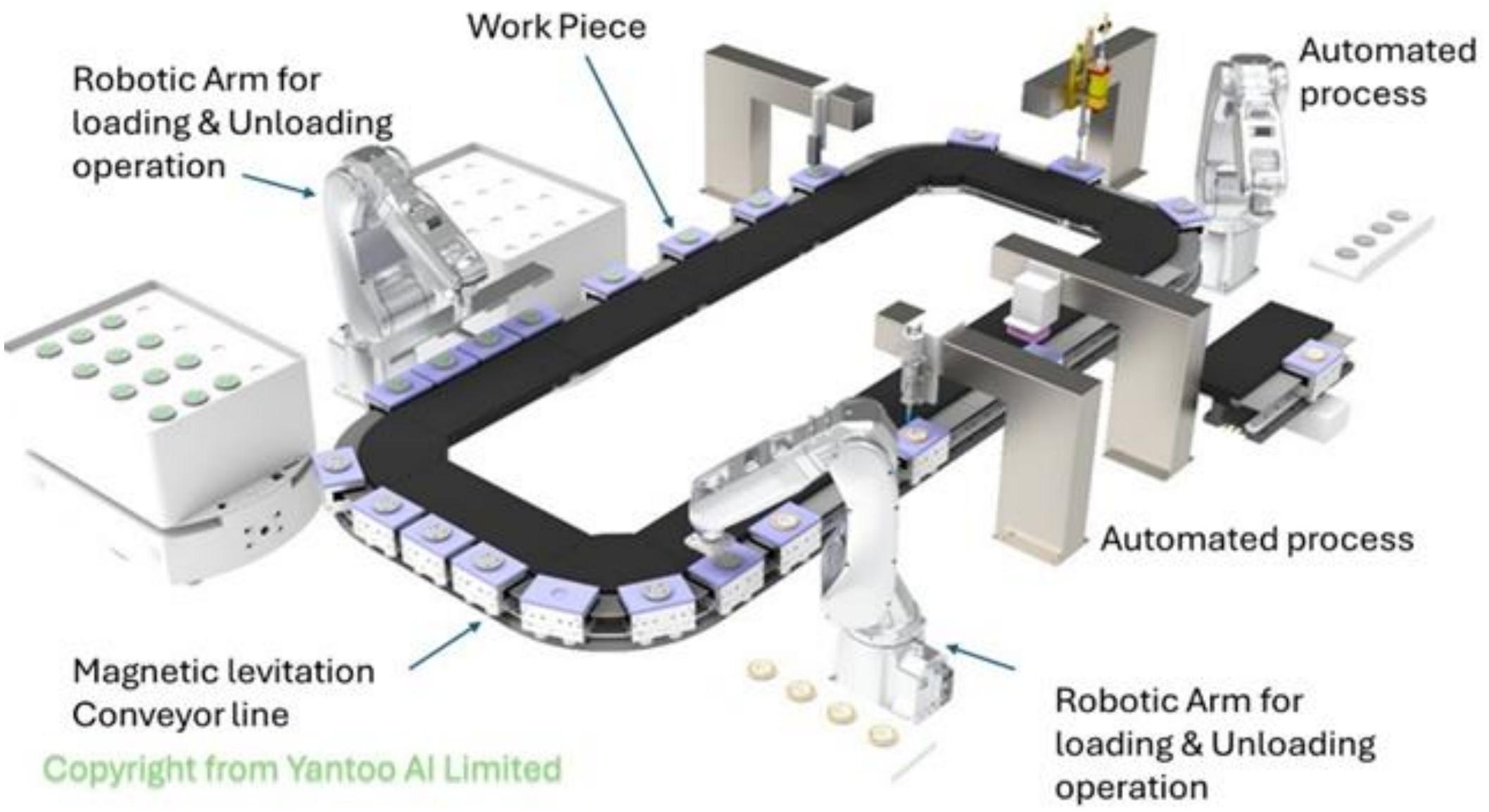

**Figure 7: Schematic Diagram of Magnetic Levitation Conveyor line with full automation in the manufacturing process**

The Magnetic Levitation (Maglev) Conveyor system has been integrated with the robots and automated machinery, as well as an integrated control system to control the whole integrated automation manufacturing process. The robotic arm for loading and uploading workpieces and components on the Maglev Mover is the same as the work-in-progress holder.

The Maglev Mover can locate the required location with high precision for an automation process. There is no need to change the holder for the assembly line.

In the design of the Magnetic levitation line, it applies the electromagnetic force and a controller to create levitation force and repulsion force for the movement of the mover at high speed and high accuracy.

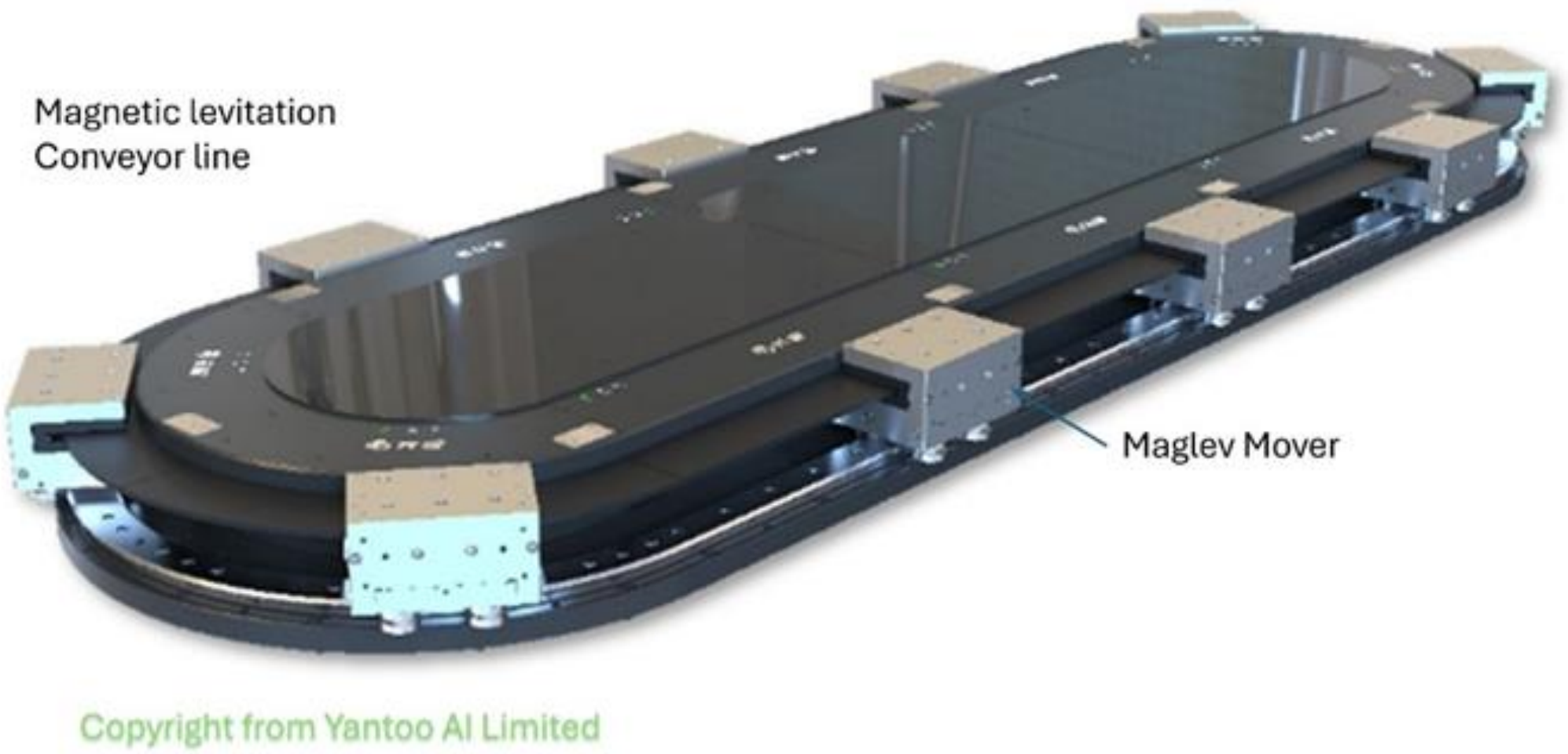

**Figure 8: Magnetic Levitation Conveyor Line Diagram from Company Y**

The Magnetic levitation Conveyor Line Diagram in Fig. 8 from Company Y has applied the permanent magnet and electromagnet to ensure the lift-up of stability of the mover by levitation force and drive the movement of the mover at high speed and high accuracy. The magnetic levitation conveyor has been designed by the company's Y engineers. During the pilot run, it can perform the lift-up stability of the mover and the speed of the mover with the integrated control. Figure 9 shows the photo of the production process in automation with a Magnetic levitation conveyor.

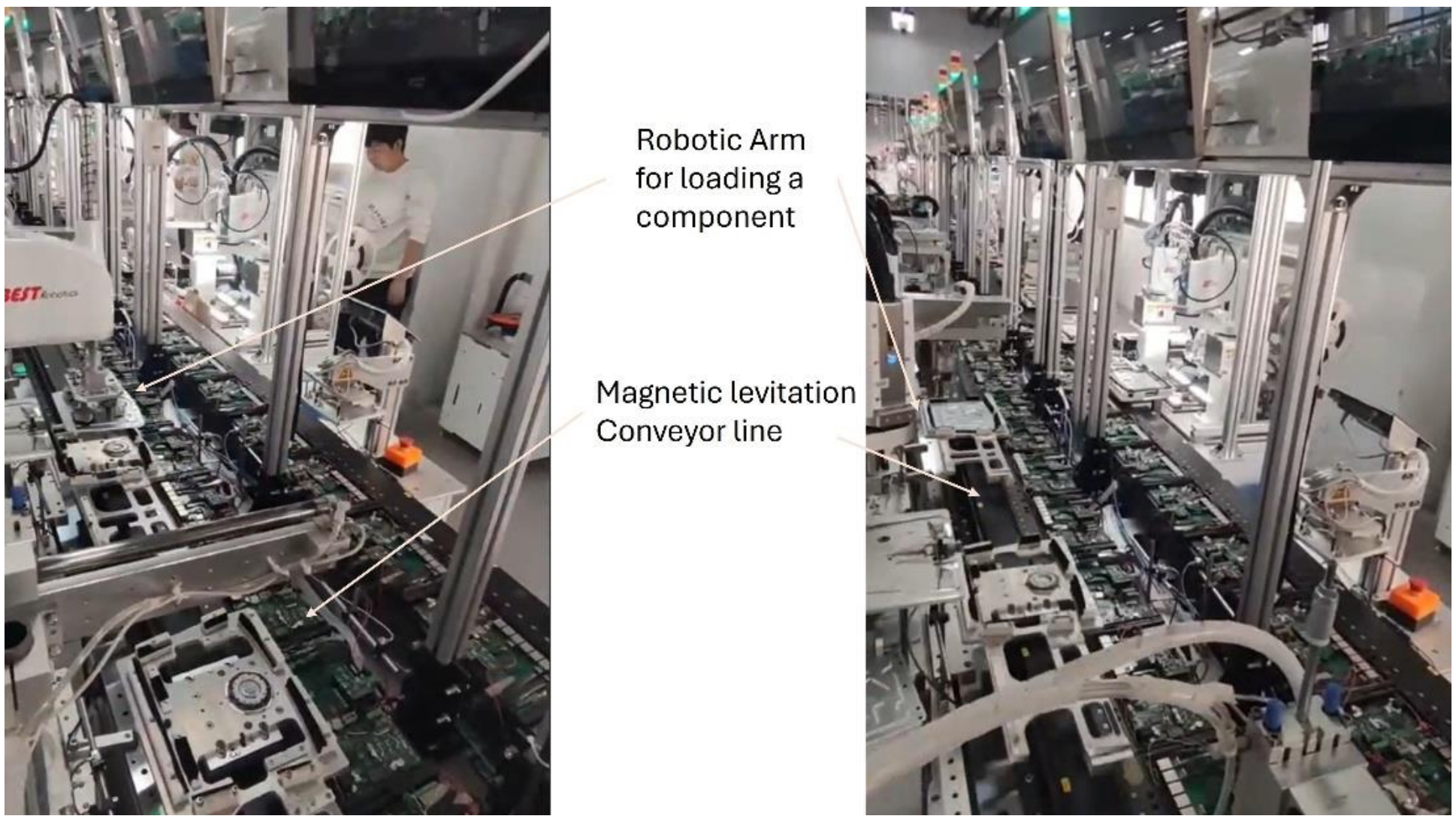

**Figure 9: Magnetic Levitation Conveyor for automated assembly in China**

## IV. METHODOLOGY AND PROJECT DEVELOPMENT

Engineering Development Plan for Magnetic Levitation Conveyor can include the following:

1. Project Overview

The objective of this project is to develop a Magnetic Levitation (Maglev) Conveyor system that utilizes permanent magnets and electromagnets to achieve high-speed, high-accuracy movement while ensuring the stability of the mover through levitation forces. This development will involve design, prototyping, testing, and implementation phases to create a reliable and efficient conveyor system suitable for automated production processes.

2. Project Phases

Phase 1: Research and Feasibility Study

Objective: Assess the technical and economic feasibility of the Maglev conveyor system.

Activities:

- Conduct a literature review on existing Maglev technologies and applications.
- Analyze market needs and potential applications in various industries.
- Evaluate the cost implications and potential return on investment (ROI).
- Identify regulatory and safety standards relevant to the design and operation of the conveyor system.

Phase 2: Conceptual Design

Objective: Develop initial design concepts for the Maglev conveyor system.

Activities:

- Create preliminary design sketches and system architecture.
- Define specifications for key components, including permanent magnets, electromagnets, control systems, and structural elements.
- Use computer-aided design (CAD) software to model the conveyor system.
- Conduct simulations to evaluate the performance of the proposed designs under various operational conditions.

Phase 3: Detailed Design and Prototyping

Objective: Finalize the design and build a prototype of the Maglev conveyor system.

Activities:

- Refine the design based on simulation results and feedback from stakeholders.
- Select materials and components for the prototype, ensuring they meet performance and safety standards.
- Fabricate the prototype, including the magnetic levitation mechanism, control systems, and structural components.
- Develop software for integrated control of the conveyor system, including speed regulation and stability monitoring.

Phase 4: Testing and Validation

Objective: Evaluate the performance of the prototype through rigorous testing.

Activities:

- Conduct initial tests to assess the lift-up stability of the mover and its speed and accuracy.
- Perform stress tests to evaluate the system's durability and reliability under various load conditions.
- Analyze data collected during testing to identify any performance issues or areas for improvement.
- Iterate on the design as necessary based on testing outcomes.

Phase 5: Implementation and Deployment

Objective: Prepare the Maglev conveyor system for commercial deployment.

Activities:

- Develop installation guidelines and operational manuals for end-users.
- Train personnel on the operation and maintenance of the Maglev conveyor system.
- Collaborate with manufacturing partners for the mass production of the system.

- Launch the conveyor system in a pilot production environment to monitor performance and gather user feedback.

Phase 6: Post-Implementation Review

Objective: Assess the success of the Maglev conveyor system and identify opportunities for future improvements.

Activities:

- Collect feedback from users regarding system performance, reliability, and ease of use.
- Analyze operational data to evaluate the impact of the Maglev conveyor on production efficiency and cost savings.
- Identify any issues that arose during implementation and propose solutions for future iterations.
- Document lessons learned and best practices for future projects.

3. Methodology

Agile Development Approach: Utilize an iterative and incremental approach to allow for flexibility and adaptability throughout the development process. Regular feedback loops with stakeholders will ensure that the project remains aligned with user needs and market demands.

- Cross-Functional Collaboration: Engage a multidisciplinary team of engineers, designers, and industry experts throughout the project to leverage diverse expertise and perspectives.
- Prototyping and Testing: Emphasize rapid prototyping and iterative testing to validate design concepts and ensure that the final product meets performance specifications.
- Data-Driven Decision Making: Utilize data analytics to inform design choices, assess performance during testing, and guide improvements throughout the development process.
- Compliance and Safety Standards: Ensure that all designs and implementations adhere to relevant industry standards and regulations to guarantee safety and reliability.

The Engineering Development Plan for the Magnetic Levitation Conveyor outlines a structured approach to designing, prototyping, and implementing a cutting-edge conveyor system. By following this plan and methodology, Company Y aims to deliver a high-performance Maglev conveyor that meets the demands of modern automated production environments while ensuring stability, speed, and efficiency.

## V. ARTIFICIAL INTELLIGENCE IN THE MAGNETIC LEVITATION CONVEYOR

Artificial Intelligence (AI) can be applied to Magnetic Levitation (Maglev) Conveyor systems in several ways to enhance their functionality and efficiency:

- Real-Time Monitoring and Control: AI algorithms can continuously monitor the conveyor system's operations, analyzing data from sensors to ensure optimal performance. This includes tracking the speed, position, and condition of goods being transported, allowing for immediate adjustments to maintain efficiency and prevent disruptions.
- Predictive Maintenance: AI can analyze historical and real-time data to predict potential maintenance issues before they occur. By identifying patterns and anomalies, AI can schedule maintenance activities proactively, reducing downtime and extending the lifespan of the conveyor system.
- Adaptive Routing: AI can dynamically adjust the routing of goods based on current production demands and system conditions. This ensures that materials are transported in the most efficient manner, optimizing throughput and minimizing bottlenecks.
- Energy Optimization: AI can optimize energy consumption by adjusting the magnetic levitation power based on the weight and size of the items being transported. This reduces energy usage and contributes to more sustainable operations.
- Quality Control: AI can integrate with optical inspection systems to perform real-time quality checks on products as they move through the conveyor. This ensures that defective items are identified and removed promptly, maintaining high-quality standards.

- Scalability and Flexibility: AI can facilitate the integration of new processes or changes in production requirements without extensive reconfiguration. By learning from past adjustments, AI can quickly adapt to new conditions, enhancing the system's flexibility.
- Data-Driven Insights: AI can analyze data collected from the conveyor system to provide insights into production efficiency, identifying areas for improvement and enabling informed decision-making.

By leveraging AI, Maglev Conveyor systems can achieve higher levels of automation, efficiency, and adaptability, making them a powerful tool in modern manufacturing environments. Prof Ray WM Kong [17] [18] has developed an AI visual recognition to apply for the Maglev Conveyor to be an AI Maglev Conveyor for coming future.

## VI. CONCLUSION

The magnetic levitation conveyor from Company Y consists of an integrated electromagnetic generator and controller, and more movers. By using the Maglev system, engineers can install the automation equipment and movers in fixed positions as an initial stage. The movers are complex, both mechanically and electrically, and the ability to move the workpieces rather than the chain conveyor or belt conveyor increases the flexibility and accuracy of the feed workpiece to the automated machinery.

Flexibility was an additional benefit. The Maglev Conveyor attach a variety of material and workpieces onto the mover for production using simple adapters as shown in the Fig. 9. We can easily add processing stations alongside the assembly of electronics products for good parts at high precision and accuracy, for instance, or optical sensing heads to conduct full part inspections—and carry workpieces to them as needed. The maglev conveyor's mover can provide a rapid acceleration, also letting us move material samples at high speeds; with thin samples, for instance, this helps minimize treatment time with the electronics product assembly.

Another benefit was cost reduction. Conventional setups use a six-axis robot or linear motor to move a robotic gripper to grasp the components of a stationary workpiece. The maglev conveyor and system can move a short distance to accommodate robotic arm loading and fit the components to the casing and PCB. The traditional conveyor cannot make a small adjustment to fit the robotic assembly requirement. From a cost perspective, the Maglev Conveyor comes in somewhere between linear-axis and robotic systems. With flat parts that don't require much vertical travel on the Z-axis, where robotic systems are usually ideal, it offers an excellent alternative to various control systems.

Furthermore, the new development of AI magnetic levitation conveyor can bring an additional benefit on top of the Magnetic Levitation Conveyor for automation.